\documentclass{ifacconf}

\makeatletter
\let\old@ssect\@ssect 
\makeatother

\usepackage{graphicx}      
\usepackage{natbib}        
\usepackage{amsmath}
\usepackage{amssymb}
\usepackage{mathtools}
\usepackage{pdfpages}
\usepackage{booktabs}

\usepackage{pgfplots}\pgfplotsset{compat=newest}
\usepgfplotslibrary{groupplots, dateplot}
\usepackage{tikz}
\usepackage{tikzscale}
\usetikzlibrary{external}
\usepackage{xurl}
\usepackage{hyperref}

\makeatletter
\def\@ssect#1#2#3#4#5#6{%
  \NR@gettitle{#6}
  \old@ssect{#1}{#2}{#3}{#4}{#5}{#6}
}
\makeatother
\begin{document}
\begin{frontmatter}

\title{Reinforcement Learning from Simulation to Real World Autonomous Driving using Digital Twin  \thanksref{footnoteinfo}} 

\thanks[footnoteinfo]{This work is part of FOCETA project that has received funding from the European Union’s Horizon 2020 research and innovation programme under grant agreement No 956123 and ELO-X No 953348.}

\author[First]{Kevin L. Voogd}  
\author[Second]{Jean Pierre Allamaa}
\author[First]{Javier Alonso-Mora}
\author[Second]{Tong Duy Son}

\address[First]{Faculty of Mechanical, Maritime and Materials Engineering, Technical
University of Delft, 2628 CD Delft, The Netherlands
 (e-mail: k.l.voogd@student.tudelft.nl, j.alonsomora@tudelft.nl).}
\address[Second]{Siemens Digital Industries Software, 
   Leuven, 3001 Belgium (e-mail: \{jean.pierre.allamaa, son.tong\}@siemens.com)}

\begin{abstract}                
Reinforcement learning (RL) is a promising solution for autonomous vehicles to deal with complex and uncertain traffic environments. The RL training process is however expensive, unsafe, and time consuming. Algorithms are often developed first in simulation and then transferred to the real world, leading to a common sim2real challenge that performance decreases when the domain changes.  In this paper, we propose a transfer learning process to minimize the gap by exploiting digital twin technology, relying on a systematic and simultaneous combination of virtual and real world data coming from vehicle dynamics and traffic scenarios. The model and testing environment are evolved from model, hardware to vehicle in the loop and proving ground testing stages, similar to standard development cycle in automotive industry. In particular, we also integrate other transfer learning techniques such as domain randomization and adaptation in each stage. The simulation and real data are gradually incorporated to accelerate and make the transfer learning process more robust. The proposed RL methodology is applied to develop a path following steering controller for an autonomous electric vehicle. After learning and deploying the real-time RL control policy on the vehicle, we obtained satisfactory and safe control performance already from the first deployment,  demonstrating the advantages of the proposed digital twin based learning process. 

\end{abstract}


\end{frontmatter}

\section{Introduction}\label{chapter:introduction}

\par Research on autonomous vehicles (AVs) has made significant progress with recent advances of deep learning (DL), especially on the vehicle perception stack. While there have been some encouraging results and demonstrations, the application of DL on the vehicle planning and control stacks are still limited. Deep reinforcement learning (DRL) is an approach to generate control strategies in sequential processes, and capable to automatically learn and adapt from data, robust to different operating conditions and tasks. This offers a more flexible and higher performance planning or control solution than traditional model-based control methods, which rely on well-defined mathematical model of the system. Recent DRL breakthroughs example include AlphaStar \citep{arulkumaran_alphastar_2019}, a model designed to play StarCraft II and end-to-end autonomous lane keeping driving \citep{kendall_learning_2018}. 

Despite the progress, further research and testing are crucial to realize the advantages of DRL, be implementable in real world and automotive industry standard \citep{you_advanced_2019}. The main DRL challenge is a safe and efficient training process and testing environment. DRL training is expensive, time consuming and involves with exploration of unsafe, risky situations. Training with physical car is not possible in real traffic legally and is often limited to closed tracks. Once may first train with collected human driving data and then deploy the controller in the physical world; however, the data misses out on critical scenarios that needed to robustify the controller. The alternative training environment is simulation, where virtually generated data is cheap and fast, already labeled, and includes a large number of critical scenarios. Still, virtual data lacks true real life properties and interactions. When a policy is trained in simulation and then deployed testing in the physical world, the performance differs, known as the {sim2real} transfer gap. This gap is caused by the simulation-optimization-bias where the controller exploits faults in the simulator and overestimates its performance compared to the target domain \citep{muratore_assessing_2021}. Other causes are model mismatch, noise, or actuation delays. 

In this paper, we propose a DRL training and testing environment relying on Digital Twin (DT). DT is a virtual representation of a physical product or process and being used across its development cycle to simulate and optimize the system's performance and efficiency. In AVs, it comprises virtual models of vehicle dynamics, traffic scenarios and sensors. We also exploit closed-loop DT which provides bi-directional connectivity between the physical and the virtual data. In particular, the fidelity and complexity of models and environment are gradually evolved from model-in-the-loop (MiL) in simulation to hardware-in-the-loop (HiL) with physical vehicle embedded controller and actuation components, and vehicle-in-the-loop (ViL) proving ground testing with a real Siemens Simrod drive-by-wire car. The process provides a robust, high performance transfer learning, and easier for prediction and tuning. Note that this is also known as V-cycle in industry, being adapted to DRL development purpose. Finally, we combine with domain adaptation and domain randomization techniques to enhance the transfer learning process. 

The transferred controller is deployed in the real vehicle successfully without fine-tuning in the target domain. The contributions of this paper are:
\begin{itemize}
	\item a zero-shot transfer learning approach that combines the advantages of virtual training with real-world data. The DRL agent is robust to different paths and model uncertainties,
	\item a reduction in the sim2real gap for autonomous driving applications. The RL agent is trained using a high-fidelity (HF) vehicle dynamics simulator and traffic scenario simulator with domain randomization and adaptation,
        \item a deployable algorithm on a real-time operating system and a validation framework in MiL, HiL and ViL minimizing the overall testing effort and cost.
\end{itemize}

\par The paper is organized as follows. Section \ref{sec:relatedwork} summarizes related work on reinforcement learning for path following and sim2real methods. Section \ref{sec:background} reviews the background theory of RL, the vehicle model, and the transformations needed in domain adaptation. The experimental setup  and the implementation details employed in this work are presented in Section \ref{sec:experiments}, Section 5 provides a discussion on the results and concludes this work.

\section{Related work}
\label{sec:relatedwork}
\subsubsection[DRL and path following in autonomous driving:]{\texorpdfstring{DRL and path following in autonomous  driving:}{DRL and path following in autonomous driving:}}
the first successful application of DRL in autonomous driving was achieved by \cite{kendall_learning_2018}, learning a control policy for lane-keeping from monocular images and training only in the physical world. Recent work by \cite{alomari_path_2021} claims to have developed a method that bridges the Sim2Real gap using a 3D vehicle dynamics simulator and parameter randomization, but the results are not validated in a real-time operating system. A similar study by \cite{maramotti_tackling_2022} focuses on a DRL planner using the single-track kinematic model and an additional neural network (NN) to simulate the state transition dynamics of the car. To speed up convergence, they pre-train the network with imitation learning and randomize the path and the vehicle's initial state. Similarly, \cite{jiang_pathfollowing} uses NNs to model the vehicle dynamics and constraint weights and activation functions of the DRL algorithm to turn it into a convex optimization problem. However, the computational time is significant, making it unfeasible to deploy in a real-time application. 

\subsubsection{Sim2Real methods:}
Domain adaptation (DA) maps features from the source domain to the target domain, and \textit{vice versa} or both, to a common latent space in an attempt to train the agent in a domain-independent framework. DA has been used to transform synthetic images or point clouds into realistic representations through generative adversarial networks \citep{pan_virtual_2017}. Other researchers argue that data representation is the main source of the transfer gap and propose to transform the representation to lidar maps \citep{wang_pseudo-lidar_2019} or bird-eye views \citep{ng_bev-seg_2020}.
Domain randomization (DR) is a method in which the parameters of the source domain are randomized so that it contains the target domain in its distribution. This method is effective when used with domain-specific knowledge and results in a robust control policy to model uncertainties as performed in \cite{allamaa_sim2real_2022} to automatically tune the controller parameters. Randomization techniques have been applied to vehicle dynamics and physical parameters such as masses, friction, \citep{peng_sim--real_2018}, trajectories, or random forces \citep{pouyanfar_roads_2019}. It has also been performed on sensor data where DR techniques to alter poses, textures, dimensions, or colors \citep{openai_learning_2019}.
Lastly, system identification is used to identify properties of dynamical systems based on experimental measurements, which are later used to simulate the process more accurately. In addition, a Digital Twin is a high-fidelity multiphysics model that uses the available models and sensors to recreate in simulation its real life counterpart. By combining this model with real-time data, it is possible to continuously predict the behavior of the vehicle in the most realistic way \citep{hartmann_digital_2020}.

We propose on a transfer learning approach to a real-time application that combines all the three techniques in a systematic way, together with using integrated high-fidelity virtual and real-world data efficiently.

\section{Background}\label{sec:background}
This section introduces the theory of RL, the vehicle kinematic model used in this work and lastly, the calculatation of the deviations from the reference path.

\subsection{Reinforcement learning}
\par Formally, RL problems are formulated as Markov decision processes (MDPs). Specifically in autonomous driving, RL is used to solve the MDP for the optimal driving policy. At every step, the MDP is composed by the tuple $(\mathcal{S,A,P},R,\gamma)$, where $\mathcal{S}$ and $\mathcal{A}$ are the set of states and actions, respectively. In stochastic processes $\mathcal{P}(s_{t+1}|s_t, a_t): \mathcal{S}\times\mathcal{A}\rightarrow [0,1]$ is the transition probability function of entering the state $s_{t+1}$ from state $s_t$ by taking the action $a_t$. Moreover, MDPs have the property that the conditional probability of a future state depends only on the present state. $R: \mathcal{S}\times\mathcal{A}\times\mathcal{S}\rightarrow\mathbb{R}$, is the reward function that maps the state $s_t$ in which the action $a_t$ is taken and the resulting state $s_{t+1}$ into a scalar value . The action is chosen based on a policy $\pi_\theta(a|s)$. In DRL, the policy is approximated with deep NNs with parameters $\theta$. Given stochastic transition and policy functions, the objective is to maximize the return $R_t$ over the trajectory, i.e. the expected reward: \begin{equation}
	J(\pi) = \int_{\tau}\mathcal{P}(\tau|\pi)R_t = \mathbb{E}\left[R_t\right].
\end{equation} The policy that maximizes the objective $J(\pi)$ is noted as $\pi^*$. In addition, there are two other functions: the state-value $V^{\pi}(a|s)$ function, which is the expected return of following the policy $\pi_\theta(a_t|s_t)$ from state $s$. The $Q^{\pi}(a|s)$ is the action-value function and maps the expected return of choosing an arbitrary action $a$ in state $s$ and then effectively follow the policy $\pi(\cdot|s)$. 
In this paper, we use the soft-actor critic (SAC) algorithm developed in \cite{haarnoja_soft_2018}, which is based on an actor-critic structure: the \textit{actor} selects the action of the agent and the \textit{critic} evaluates the action by approximating the $Q_\pi$-function. SAC includes entropy regularization in its objective function to encourage exploration, preventing early convergence to bad local minima.  Additionally, this algorithm is off-policy meaning that the $Q_\pi$-function is learned from actions taken by a different policy than the current $\pi(a|s)$. 

\subsection{Single-track model}
\begin{figure}[tb]
	\centering
	\includegraphics[width=0.75\linewidth]{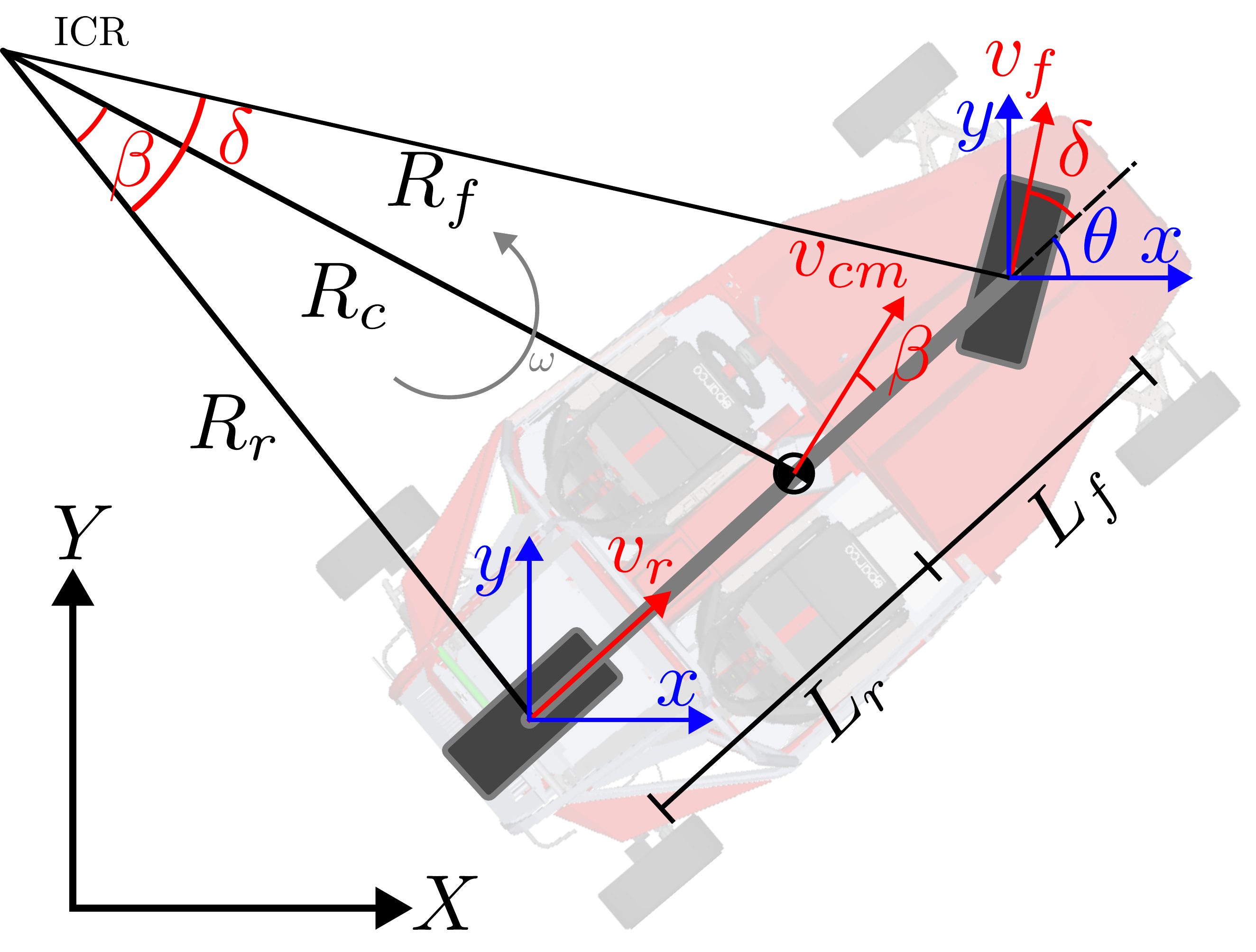} 
	\caption{Single-track model}
	\label{fig:bicycle_model}
\end{figure}

\par The single-track bicycle model is a kinematic model of a four-wheeled vehicle (Fig. \ref{fig:bicycle_model}), in which the wheels at each axle are joined together. This model assumes a no-wheel slip condition. The length of the wheelbase is denoted as $L$, and the distances from the rear and front axle to the vehicle's center of gravity (CoG) are $L_r$ and $L_f$, respectively. The vehicle's linear and angular velocities measured in a global reference frame are:
\begin{align}
	\label{eq:x_dot}
	\dot{x}_{cm} &= v\cos{\left(\theta +\beta\right)} \nonumber\\
	\dot{y}_{cm} &= v\sin{\left(\theta +\beta\right)}\\
	\dot{\theta} &= {v}/{R_{c}}, \nonumber
\end{align}
where $\theta$ is the heading of the chassis with respect to the global frame and $\beta$ is the angle enclosed by $v_{cm}$ and $v$. With $\beta=\arctan\left(({L_r}/{L})\tan\delta\right)$ and the CoG's radius of rotation is  $R_c={L}/({\tan\delta\cos\beta})$, the state variables can be described in terms of the inputs $\delta$ and $v$. 

\subsubsection{Lateral and deviation:}
\begin{figure}[tb]
	\centering
	\includegraphics[width=.75\linewidth]{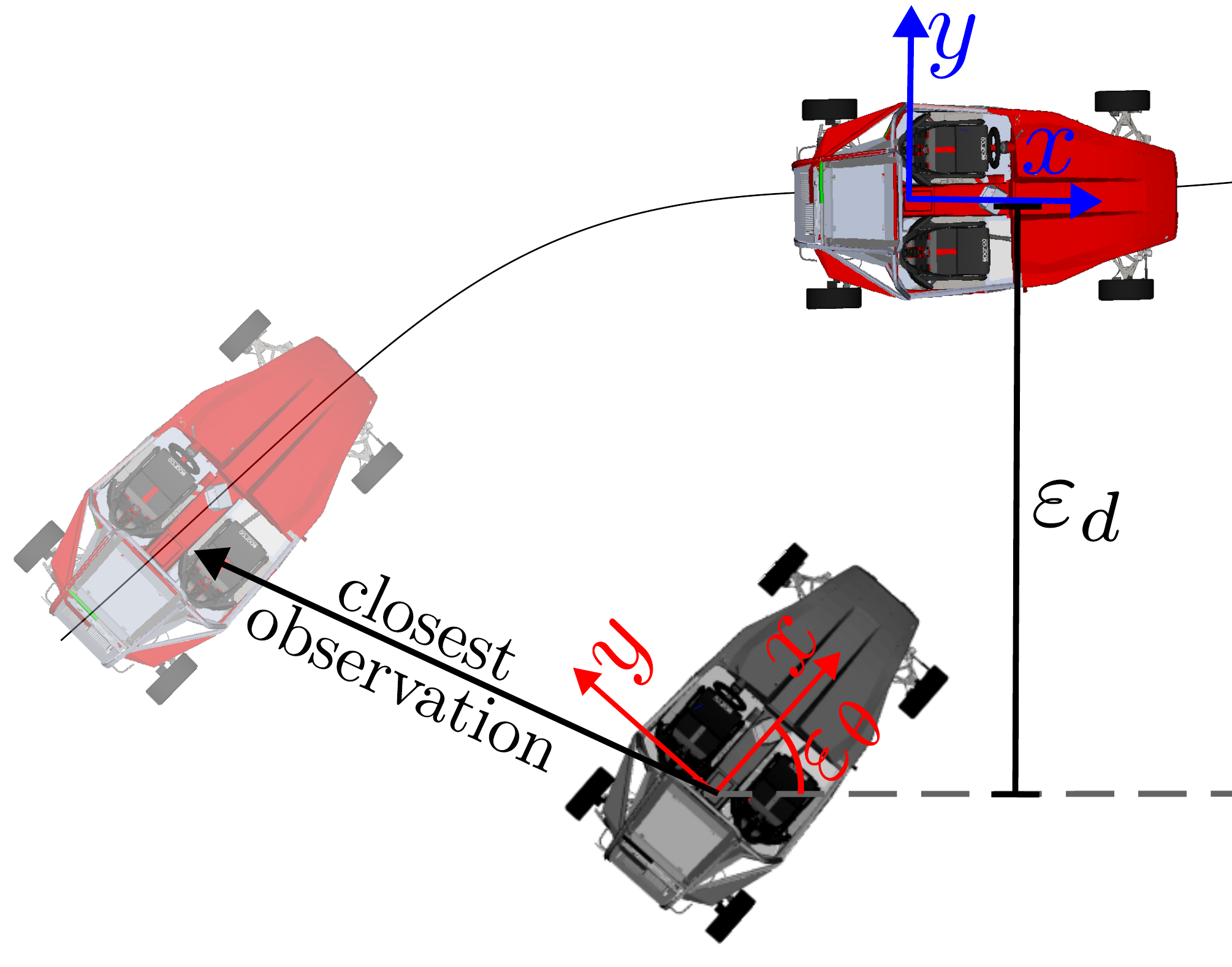}
	\caption{Lateral deviation and yaw error calculation for the learning agent (gray). The deviation is computed w.r.t. the closest next observation (red).}
\label{fig:lateral_dev}
\end{figure}

In this work, we use a buffer of recorded virtual and real-world trajectories to represent the centerline of the path to be driven. We then calculate the heading and lateral deviations with respect to the closest next point in distance in such buffer. In Figure~\ref{fig:lateral_dev}, the black vehicle represents the RL agent, and the red ones the logged data. The heading ($\varepsilon_\theta$) is calculated by computing the shortest difference between angles. The lateral deviation ($\varepsilon_d$) is calculated in~\eqref{eq:lateraldev} as: \begin{equation}
\label{eq:lateraldev}
	\begin{bmatrix}	dx & dy & 1 \end{bmatrix}^\text{T} = {\left(\prescript{E}{W}{\textit{H}}\right)}^{-1}\begin{bmatrix} \prescript{W}{}{x_S} & \prescript{W}{}{y_S} & 1 \end{bmatrix}^T,
\end{equation} where $S$ and $E$ are the subscripts for the learning agent and the closest next observation respectively, $W$ is the global frame, $dx$ is the longitudinal offset, $dy = \varepsilon_d$ is the lateral deviation, and $H$ is the homogeneous transformation matrix. 

\section{DRL Training and Transfer Learning}
\begin{figure*}[tb]
	\centering
	\includegraphics[width=0.9\textwidth]{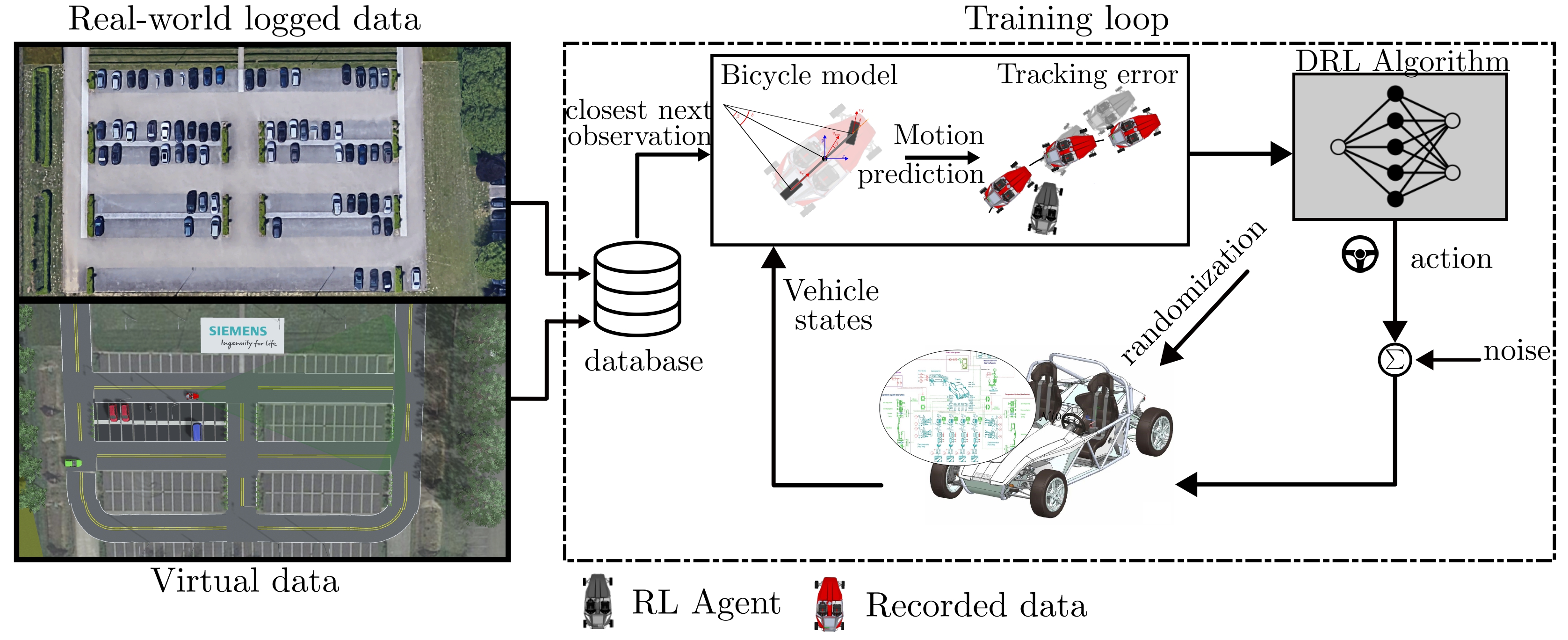}
	\caption{Training of reinforcement learning policies using first synthetically generated data and a digital twin of the vehicle until performance settles. Then,  real-world logged data is used. The predicted deviation is calculated with the single-track model. The initial states, the vehicle's physical parameters, and the control action are randomized.}
	\label{fig:pipeline}
\end{figure*}

This section starts with a description of the training loop depicted in Figure \ref{fig:pipeline} followed by a detailed explanation. First, different trajectories are recorded in simulation (Simcenter Prescan) and in the physical world and saved in the buffer. Training starts with the RL implementations \citep{stable-baselines3} with simulated data. When the performance stops improving, real-world trajectories are included as the first step in DR to generalize on the noise level and dynamic driving style. The data is set in the error frame presented in Fig.~\ref{fig:lateral_dev} as an attempt for DA. The episodes start with a random initialization of the environment and vehicle states, the second component of DR. The output of the DRL algorithm, the control action, is sent to the 15 DoF high fidelity (HF) vehicle dynamics simulator. We use the digital twin of the Simrod available on Simcenter Amesim. The HF model includes a  number of identified parameters that are also randomized, to account for the sim2real gap. This third level of DR allows for even more generalization and robustification in the transfer learning approach. The performance during training is evaluated regularly every 2500 steps in 4 randomly selected scenarios and random initial conditions. The metric used to evaluate is the average timestep reward.

\subsubsection{Data generation:}the virtual data was generated in the Simcenter Prescan traffic scenario simulator at a frequency of 20 Hz with the same format as the real-world samples to facilitate the transition. The real-world samples were collected from a drive with the physical car Simrod in the company parking lot seen in Fig.~\ref{fig:pipeline}. The data were collected with the high accuracy dGPS and stored in rosbags in the local Cartesian coordinate system.


\subsection{Reinforcement learning training setup}
\subsubsection{State space:}
The NN inputs are processed sensor readings: the position in a global reference frame and inertial measurements or virtually generated. These are the longitudinal speed of the vehicle $v_x$, the heading error $\varepsilon_\theta$, the lateral deviation $\varepsilon_d$ and their derivatives. The estimated lateral deviation with the single-track model for the next 10 timesteps is included, assuming that the speed and steering remain constant. In addition, the previous steering rate $\dot{\delta}$ and  angle $\delta$ are fed too. By providing the NN with states in the error frame, we benefit from the possibility to generalize any reference path with any center of the coordinate system. This allows training independently of the domain and task to be performed. 

\subsubsection{Action space:}
The action generated by the policy network is the steering rate $\dot{\delta}$ saturated in the range $[-0.18, 0.18]$ rad/s. We opt for a generalized state-dependent exploration (gSDE), where the noise is dependent on the state of the car for the entire duration of the episode \citep{raffin_smooth_2021}. The steering rate is then integrated to obtain the steering angle. The longitudinal acceleration is computed with a PD controller.

\subsubsection{Reward function:}
the reward function depends on the heading error $\varepsilon_\theta$, the lateral deviation $\varepsilon_d$, and  the steering rate $\dot{\delta}$. We propose a multiplication of the individual components and normalize them between 0 and 1 as shown next: 
\begin{equation}
    r = \left(1-\frac{\left|\varepsilon_\theta\right|}{\varepsilon_{\theta_\text{max}}}\right)\left(1-\frac{|\varepsilon_d|}{\varepsilon_{d_\text{max}}}\right)\left(1-|\dot{\delta}|\right).  
\end{equation}

\subsubsection{Parameter randomization:}
We randomize the digital twin physical parameters to robustify against modeling errors and uncertainties such as changing road conditions, the number of passengers, or delays. Consequently, we randomize the mass, the location of the center of gravity, the length of the wheelbase, the suspension, and the stiffness of the tire. Values are drawn from a normal distribution. In addition. the initial deviation with respect to the path is changed every episode. 

\subsection{Implementation details}
The DRL model is trained on a laptop with 64 GB of RAM, an Intel Xeon W-11855M processor, and an NVIDIA RTX A4000 laptop graphics card. The model used for training has 16 inputs, 6 layers, and only one output, and also has a linear decay on the learning rate. The preprocessing steps and the NN are C code generated and deployed to the embedded platform dSPACE MicroAutobox III, running the real-time controller that commands the SimRod to perform real-time control actions on the Simrod.
\subsection{Experiments}
\label{sec:experiments}
We train four different policies and then evaluate them in a standard V-cycle procedure satisfying the safety requirements: Model-in-the-loop (MiL), hardware-in-the-loop (HiL), and vehicle-in-the-Loop (ViL). The evaluated policies are SAC-ST-RW, SAC-HF-VD, and SAC-HF-RW, which abide by the notation: trained only with virtual data (VD), fine-tuned with real-world data (RW), single-track model (ST), and high-fidelity model (HF). The SAC-ST-RW is evaluated with the higher-fidelity model. MiL allows for safe and extensively verify and validate the trained policies against possible edge cases, and actuator noise levels and provides initial performance metrics. This step enables a safe and cheaper transition to HiL and ViL levels.
The learned policies are evaluated in real-world scenarios and are consistently evaluated at different initial deviations: $\{-1.25, -1.0, -0.5, 0.0, 0.5, 1.0, 1.5\}$ meters but the vehicle's physical parameters are kept constant to compare the performance under equal conditions. ViL is carried out in a closed parking lot in Leuven, Belgium.

\section{Results and discussion}\label{sec:results}
\begin{figure}[tb]
    \centering
    \includegraphics[width=\linewidth]{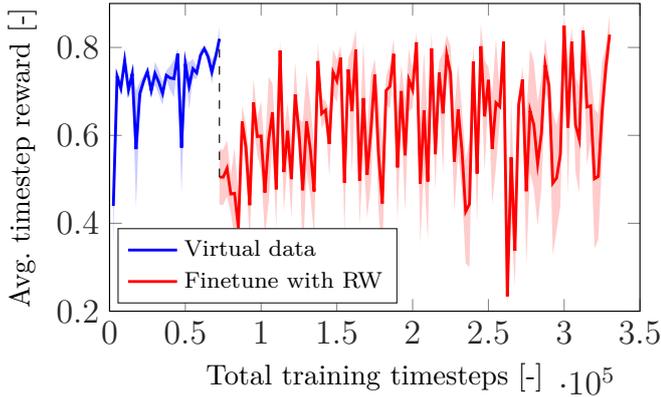}%
    \caption{Training results for SAC-HF agent: the agent is first trained with virtual data and the best model is further fine-tuned with real-world data.}
    \label{fig:training_rewards}
\end{figure}

In this section, we show the training results and present the evaluation of the MiL and ViL experiments. Figure \ref{fig:training_rewards} shows the average timestep reward of the SAC-HF-RW model, the best performing policy. The blue line shows the performance over time using only virtual data. Then, it is fine tuned with real-world data (red) as a component of the transfer learning methodology. There is a large decay (sim2real gap) when the data type is changed, showing the necessity of introducing such transfer learning logic, as the pre-trained model would have suffered from the sim2real gap. The model continues to train until settling in terms of performance. The model with the higher average return is evaluated in MiL and ViL. 
We observe a smaller variance during pretraining because the virtual scenarios are noise-free and simpler. These paths can be generated in millions and can speed up the pre-training phase, however, these are kinematic trajectories, which sometimes are not feasible for agents to track. Adding the real-world data afterward introduces dynamically possible curvature changes and state transitions. 
The amount of time required to perform 100000 timesteps with the high-fidelity model is 11.73 $\pm$ 0.35 hours. On the contrary, the kinematic bicycle model only requires 42.47 $\pm$ 0.13 minutes for the same number of steps.

\begin{figure}
    \centering
    \includegraphics[width=\linewidth]{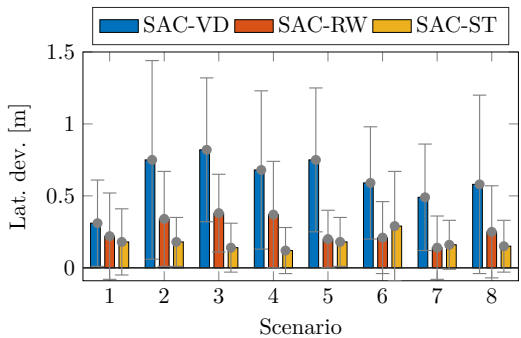}
    \caption{MiL tracking performance of the DRL policies evaluated in eight different scenarios.}
    \label{fig:MiLdev}
\end{figure}


\subsection{Model-in-the-Loop and Hardware-in-the-Loop}

\begin{figure}
    \centering
    \includegraphics[width=0.85\linewidth]{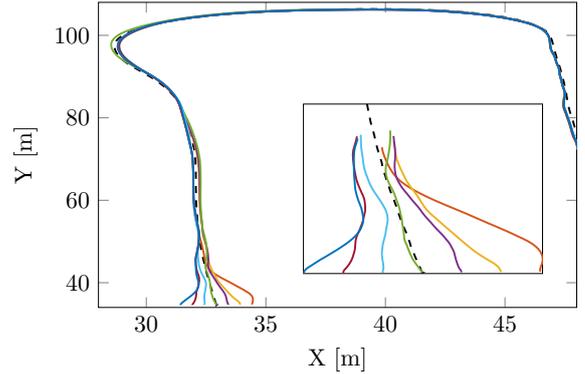}
    \caption{MiL evaluations for the SAC-ST-RW agent with different initial deviations}
    \label{fig:mil_7}
\end{figure}
The results of the MiL experiments are shown in Fig. \ref{fig:MiLdev}. In general, the SAC-ST model performs the best with an average error of 17.5 cm for all trajectories. The SAC-HF-RW model error is 26.3 cm, and then SAC-HF-VD with 62.1 cm. 
Since training scenarios are performed only in 2D and not in 3D, the HF model adds unnecessary complexity. In other words, since in MiL the target domain is still a simulator with predefined parameters, the policy tends to compromise performance for robustness. This is also the reason for using DR, as the learning agent tends to overtune in simulation, creating an SOB when transferring to the real-target domain.
However, the ST performs worse than the HF model when dealing with a set of curves or higher-speed scenarios. The former is related to the fact that the executed steering angle is subjected to the assumption of a small steering motion and a large radius of curvature. We also noticed that using only virtual data with a HF model (SAC-HF-VD) is not enough for a performant transfer, hence the benefit of our proposed transfer learning approach. 

\subsection{Vehicle-in-the-Loop}
\begin{figure}
    \centering
    \includegraphics[width=0.81\linewidth]{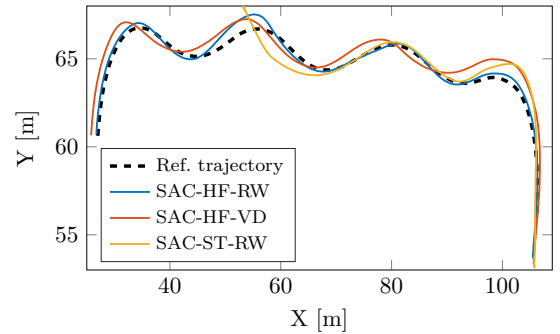}
    \caption{ViL evaluation in the same path for all agents}
    \label{fig:vil_7}
\end{figure}

As the main objective of this work is to transfer the policy efficiently and in zero-shot between the domains, we validate it in the actual vehicle. 
To our surprise, the SAC-ST-RW decreases its performance considerably when transferred to the real target. Hence, the benefit of the safe transfer learning methodology is proven, as the SAC-ST-RW does not account for a sim2real gap. The results are shown in Table \ref{tab:vil}, pointing out that the best model is the policy trained with the HF model with an average deviation of 35 cm for all paths. On the contrary, the average error for the ST model increases to 49 cm; for SAC-HF-VD, the error is 52.5 cm.  We also compare the transfer gap by computing the ratio between the performance in ViL and MiL. The SAC-ST-RW model deviates 4.75, 1.55, and 2.81 times more in the paths where ViL was tested. In contrast, the performance of SAC-HF-RW is not affected to the same extent by the sim2real gap. Specifically, we observe the following ratios: 1.60, 1.05, and 1.50. The HF model increases the tracking accuracy on average by 28.6\% compared to the single-track model. Moreover, using real-world data to fine-tune the model results in 33.3\% better transfer. Contrary to what has been reported by \cite{truong_rethinking_2022}, our results indicate that a higher-fidelity model is indeed necessary to minimize the reality gap. There were random locking of the vehicle's steering wheel in various trials; nevertheless, the agent was able to recover from the faulty states in the majority of cases and follow the path immediately afterward. 

\begin{table}[tb]
\caption{ViL results of the DRL policies tested in different scenarios}
\label{tab:vil}
\resizebox{\columnwidth}{!}{%
\begin{tabular}{lllll}
\hline
          & \multicolumn{4}{c}{Lateral deviation ($\mu \pm \sigma$) [m]} \\ \hline
 & \multicolumn{1}{c}{Path 3} & \multicolumn{1}{c}{Path 6} & \multicolumn{1}{c}{Path 7} & \multicolumn{1}{c}{Path 8} \\ \hline
SAC-HF-RW & 0.59+0.41     & 0.22+0.19     & 0.21+0.21     & 0.39+0.43    \\
SAC-HF-VD & 0.50+0.55     & 0.62+0.34     & 0.47+0.40     & 0.51+0.53    \\
SAC-ST-RW & 0.57+0.33     & 0.45+0.39     & 0.45+0.48     & -            \\\hline
\end{tabular}%
}
\end{table}
\section{Conclusion}
\label{sec:conclusion}

In this work, we develop a transfer learning strategy to efficiently train a DRL policy in simulation and deploy it in a real-time vehicle application. We show that standard approaches of training exclusively with virtual data or low-fidelity models are not sufficient to robustify the trained agent, even though they yield better performance in MiL. We combine state-of-the-art sim2real methods such as DR, DA, and HF with virtual and real-world data and show that they are all necessary components for safe transfer. 
The HF dynamics simulator allows efficient randomization of a large variety of parameters and correctly predicts the behavior of the vehicle under different conditions, robustifying the controller to real-world conditions and allowing a better zero-shot transfer. 
This work also focused on a safe and scalable approach to prototyping and developing algorithms for autonomous driving applications with physical testing in automotive industry standard. Finally, we validate our approach on a real-time path following control application in MiL, HiL, and ViL development stages. 

\bibliography{ifacconf}                                              

\end{document}